\documentclass[11pt]{article}
\pdfoutput=1 % suggested in arXiv website
\usepackage{amsmath, bm, amssymb}
\usepackage[margin=1.0in]{geometry}
\usepackage[square]{natbib}
\usepackage{graphicx} % To include figures
\usepackage{authblk} % To allow affiliations
\usepackage[section]{placeins} % Keep floats in their own section
\usepackage[textfont=small, labelfont=bf]{caption}

\def\mathbi#1{\textbf{\em #1}}

\def\BibTeX{{\rm B\kern-.05em{\sc i\kern-.025em b}\kern-.08emT\kern-.1667em\lwer.7ex\hbox{E}\kern-.125emX}}

\title{Simulation of neural function in an artificial Hebbian network}

\author{J. Campbell Scott$^1$, Thomas F. Hayes$^{1,2}$, Ahmet S. Ozcan$^1$ \& Winfried W. Wilcke$^1$}
\date{%
    \begin{flushleft}
    $^1$IBM Research, Almaden, San Jose, California, USA%
    \newline
    $^2$Univ. of Chicago, Chicago, Illinois, USA%
    \end{flushleft}
}

\begin{document}
\bibliographystyle{plainnat}
\maketitle

\begin{abstract}
Artificial neural networks have diverged far from their early inspiration in neurology.
In spite of their technological and commercial success, they have several shortcomings, most notably the need for a large number of training examples and the resulting computation resources required for iterative learning.
Here we describe an approach to neurological network simulation, both architectural and algorithmic, that adheres more closely to established biological principles and overcomes some of the shortcomings of conventional networks.
\end{abstract}

The mammalian brain is a remarkable organ. Among many other functions, it receives and interprets sensory information, plans courses of action based on both recent observations and memories of past events, then controls and coordinates muscle movements to achieve a chosen goal.
This process of sensorimotor control allows the animal to eat and to avoid being eaten, using power of about 10 Watts.

As a result of recent improvements in computer hardware and advances in the curation of large databases, artificial neural networks (ANNs) are currently driving a major surge in the performance of machine learning (ML).  However, even a small server running ML jobs requires several hundred Watts\citep{Zong:2017}, and the machine that won AlphaGo\citep{Silver:2016} consumed about 1 MW\citep{Mattheij:2016}.  A large fraction of the electrical power in ML, and deep learning (DL) in particular, goes to implement the algorithm that corrects errors by propagating them backwards through the network\citep{Werbos:1982, LeCun:2015}. Convergence to minimize the cost function requires many examples and multiple passes through the training dataset. Approaches to mitigate the power burden include the use of GPUs\citep{LeCun:2015} and the development of analog synapses\citep{Burr:2017}.  An alternative approach employs dedicated systems with optimized hardware accelerators\citep{Davies:2018, Thakur:2018, YYan:2019, Wunderlich:2019}, taking many years and many people to design, build, test and deploy.

In this paper, we pursue an approach that exploits the computational power of general-purpose modern computers, while avoiding the cost of long training times on large datasets. We address an open-ended question: ``What aspects of established neuroscience are required to emulate various capabilities of a brain?''  As a first step, we describe a system that learns unsupervised, continuously, and in real time to predict multi-modal sequences and to create reduced representations of such sequences.  To begin, we identify the fundamental principles behind the design of the network architecture and the algorithms that dictate its behavior. The overriding principle is to be guided by neurology. However, since the science of human cognition is extremely complicated and far from settled, we also attempt to avoid unnecessary complications, in the spirit of ``as simple as possible, but not simpler,'' adding additional features, mechanisms and parameters only as they seem necessary.  The resulting network architecture, and the algorithms that run it is called, for reasons that will become clear, CAL (Context Aware Learning).

In all species, inspiring current ANNs used in DL, the ``brain'' is a network of neurons interconnected through axons, synapses and dendrites.   However, biological brains are much more structured and much less homogeneous than ANNs designed to date\citep{Goodfellow:2016}.  Following biological example, CAL is organized on four different length scales (see Methods and Fig. \ref{fig:Scales_4}), starting at the cellular scale of complex individual neurons.  Several neurons interconnect to form elementary neural circuits corresponding to cortical mini-columns\citep{Mountcastle:1957, Mountcastle:1997}.  In the cortex, assemblies of mini-columns interact with each other in a ``region''. At the largest scale is a hierarchical network of interconnected regions with axonal feed-forward and feedback of neural activity.

The cortex distinguishes mammals from other species.  The cortex itself has sub-structures: there are at least 5 distinct cortical layers, identified by the type, density and connectivity of their neurons\citep{Purves:2018}, p. 627.  Running perpendicular to the layers are the aforementioned mini-columns of neurons\citep{Mountcastle:1957, Mountcastle:1997}.  Various areas of the cortical sheet connect to other subcortical structures, notably the hippocampus, thalamus, basal ganglia and amygdala, which interact with the cortex to perform specific functions. It has long been recognized that the different areas of the cortex identified anatomically\citep{Brodmann:1905} perform diverse functions ranging from primitive visual sensing to language and communication, as determined by pathology\citep{Broca:1861}, EEG\citep{Hubel:1959} and/or functional MRI\citep{Guy:1999}.   These areas are connected in a hierarchy\citep{Felleman:1991, Yildirim:2019} traversed by multiple neural pathways leading to recognition\citep{DiCarlo:2012, Haile:2019}, decision\citep{Sakagami:2006, ODoherty:2017} and action\citep{Caiani:2017}. The initial design of CAL simulates, as much as practical, parts of the cortex and thalamus. It exhibits behavior that includes correlation and association of multiple inputs, memory of temporal sequences and recognition, while allowing for future inclusion of other capabilities.

Modern neuroscience teaches us to extend the McCulloch-Pitts \citep{MCP:1943} concept of the neuron as a ``point'' at which input activity from other neurons is integrated to cause firing or not. In the current view, it is the segments and branches of dendritic trees that perform the multiply-and-accumulate function\citep{London:2005}.  Synapses are highly plastic not only in weight, but also in their very existence as new ones are formed and old ones are pruned in response to neural activity, leading to changes in the topology of the neural network. Long term retention of synaptic strength is responsible for learning and memory\citep{Purves:2018} p. 172.  In CAL, updates in synaptic weight are made according to the rule first postulated by Donald Hebb\citep{Hebb:1949}: if the pre-synaptic and post-synaptic neurons fire in causal order, the synapse is strengthened. Hebb's rule has been modified to account better for relative timing\citep{Markram:1997, Shouval:2010} by introducing anti-Hebbian weakening\citep{Oja:1982, Sanger:1989, Pehlevan:2018}. Unlike the back-propagation algorithm\citep{Werbos:1982, LeCun:2015}, Hebbian learning is entirely local, involving only the activity of the axon and dendrite connected by a single synapse.

The firing of a neuron is a transient electrochemical event generating a voltage pulse (action potential) that propagates along the axon to one or more synapses where neurotransmitters stimulate ion migration through the cell membrane and depolarize the dendrite. For the time being, we ignore the dynamics of this process \citep{Purves:2018}, p. 33 and, in the spirit of simplicity, represent neural activation as a binary value determined by whether or not the neuron fires during a short time interval. This is a reasonable first approximation if the time-step is short compared to the refractory period.  It allows future extension to spike-rate based encoding, or to event-driven activation.

There is considerable experimental evidence that neural firing is sparse\citep{Olshausen:1996} in species across the animal kingdom\citep{Barth:2012}, from insects\citep{Perez-Orive:2002} to mammals\citep{Azafar:2018}. The evolution of the neocortex not only incorporates the biological advantages of sparse encoding\citep{Laughlin:2003}, but also provides computational advantages in terms of capacity, robustness\citep{Kanerva:1988} and potentially symbolic manipulation\citep{Gosmann:2019}.  CAL imposes sparse neural activity by means of a ``$k$-winners take all'' algorithm\citep{Rinkus:2010, OReilly:2013}, that simulates neural competition and inhibition\citep{DSouza:2017, Riesenhuber:1999}.  These mechanisms are viewed as integral to the learning process\citep{Maass:2000, Heeger:2017}.

The combination of hierarchy, sparse neural firing, competition and inhibition leads inevitably to a scenario\citep{Cappe:2012, Meijer:2019} of neural processes as the propagation of firing patterns among multiple, sparsely distributed active neurons along connected, diverging and converging pathways through the network. In this view, the firing patterns are data representations having semantic meaning within the context of the specific pathway and specific locus on that pathway\citep{Kanerva:1988, Plate:1995, Hawkins:2004, Rinkus:2010}. Memory is encoded in synaptic weights that are generated by previous firing patterns, and accessed by current ones.

The final principle guiding the design of CAL is that learning in the cortex takes place through time and without supervision\citep{Heeger:2017}. The process is simulated by a recurrent neural network\citep{Elman:1990, Rinkus:1993, Hawkins:2004, OReilly:2014} in which learning occurs via prediction of future input, based on the context of prior observations;  if a prediction is correct, i.e. verified by the next (ground-truth) observation, the relevant synapses are strengthened, and \textit{vice versa}.

\section*{Results.}
The numerical experiments which we describe in this section are designed to demonstrate the basic capabilities of CAL.  They employ relatively small networks in order to have short running times and therefore are not expected to show the ultimate limits of network design and algorithms.

We describe ``unit-tests'' of each component part of the network in a sequence which follows the flow of data, starting with a single region on the lowest (sensory) level of the hierarchy and scaling to multiple regions on multiple levels.  When a sensory region has a single input channel with data in the form of a time-series (stream) of scalar values, there is nothing with which to correlate, and synapse updates can be disabled: the correlator is ``hard-wired.''  At each time-step the scalar value is encoded as a binary vector that is fed forward via the axons of the binary correlator.  Each output neuron of the correlator activates, or not, a single mini-column. If an active mini-column was predicted, i.e. one or more neurons in L-II/III were predicted, these verified neurons fire, and their lateral axons carry that information throughout the layer to generate the next prediction.  This activity is also fed to the next level of the hierarchy where temporal pooling generates a representation of the ``sequence of sequences.''

Additional details of network operations are described in the Methods section.

\subsection*{Reconstruction.}
To provide a means of assessing prediction accuracy using familiar metrics, for example percentage correct or root-mean-square (RMS) error, it is necessary to interpret the prediction. Ultimately it may be desirable to provide CAL with a separate synapse array that learns how to convert (predicted) column activity to expected input, but at this early point in development we merely use the correlator ``run in reverse'' to reconstruct the predicted input. In order to ensure that the reconstruction process does not limit prediction accuracy, the algorithm is tested by comparing known input with the reconstruction of correlator output. The reconstruction process has been tested for integer, real number as well as binary image input. For this test, sequence memory is unnecessary and is disabled. The network is therefore simply the binary correlator array, with a single encoder/decoder channel.

For integers, proper operation requires that the difference between scalar input and decoded reconstruction be identically zero for every possible value in the encoder range. This is readily achieved when the correlator network is initialized with synapses of weight unity, sparsely and evenly distributed among both axons and dendrites and with synapse updates (i.e. training) disabled. When the correlator is initialized without any connections and is being trained with two or more channels, reconstruction improves rapidly as synapses are created and can reach 100\% accuracy for evenly distributed data.  However, values which occur infrequently may require many more iterations.

For a single channel of real numbers, from a random number generator in these tests, synapse updates are again disabled.  In this case, digitization errors arise because each input is rounded to the nearest encoded value.  The RMS error should initially be comparable to the resolution, $r$, of encoding and eventually converge to $r/\sqrt{12}$. The numerical factor results from averaging over the saw-tooth error distribution of rounding error, from  $-r$/2 to $+r$/2. In practise, it takes many iterations to achieve this convergence and requires a uniform distribution of values across every digitization bin. As with integers, reconstruction accuracy with multiple channels depends on the statistics of the data.

\subsection*{Correlation}
\begin{figure}[!b]
  \centering
  \includegraphics[width=10cm]{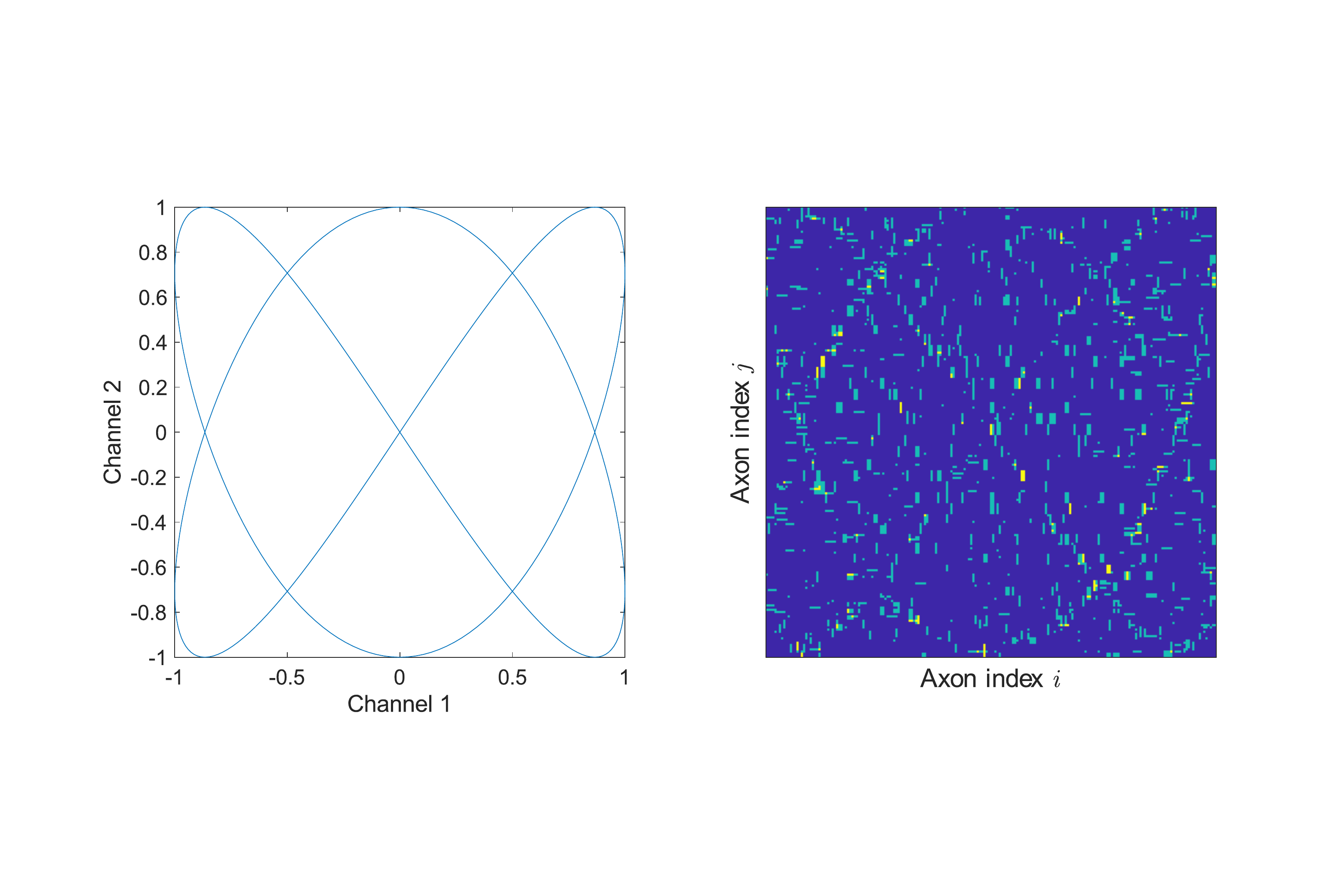}
  \caption{Demonstration of correlation.  Left: two input channels form a Lissajous figure.  Right: off diagonal quadrant of the axon weight covariance matrix, $\bm{W}^\mathsf{T}\bm{W}$, showing which axons pairs connect to common dendrites. The color scale indicates the number of shared dendrites.}
\label{fig:Lissajous}
\end{figure}
To demonstrate and test correlation, two related time-series are fed, in parallel, into a single region, with only the binary correlator active and updating its synapses.  Sequence memory is unneeded.
A suitable pair of signals are sine waves with frequencies rationally related, and with a constant phase shift. In this example,
\begin{equation}
 \label{eq:Lissajous}
s_1(t)=\sin(4\pi t/360) \textnormal{ and } s_2(t)=\sin(6\pi t/360+\pi/6)
\end{equation}
i.e. the frequencies are in the ratio 2:3, the phase shift is $\pi$/6 and the pattern repeats every 360 time-steps.  The two input channels were encoded as binary vectors of length 205 bits each, of which 5 were active.  Since the range of the data is [-1, 1] and there are 201 encoding steps (''bins'') the resolution is 0.01.  The output vector of the correlator had length 1024, with $k=32$ bits activated.

These network parameters result in an initial RMS reconstruction error of approximately 0.004, only slightly greater than the 0.003 expected for uniformly distributed inputs.  As synapses are connected and disconnected in response to the correlation in input, the reconstruction error increases slightly, leveling off at 0.01 after 100,000 iterations.

Fig. \ref{fig:Lissajous} illustrates the response of the synapses connections to correlated input. It shows the lower left quadrant of the connectivity covariance matrix $\bm{W}^{\mathsf{T}}\bm{W}$, where $\bm{W}$ is the 1024 x 410 correlator weight matrix.  Each pixel $(i, j)$ in the image is non-zero if axons $i$ and $j$ connect to same dendrite.  It is apparent that axons 1 - 205 (signal $s_1$) are correlated with axons 206 - 410 (signal $s_2$) according to a Lissajous figure of ratio 2/3. The correlator is indeed performing as expected.

\subsection*{Sequence memory.}
\begin{figure}[!b]
  \centering
  \includegraphics[width=12cm]{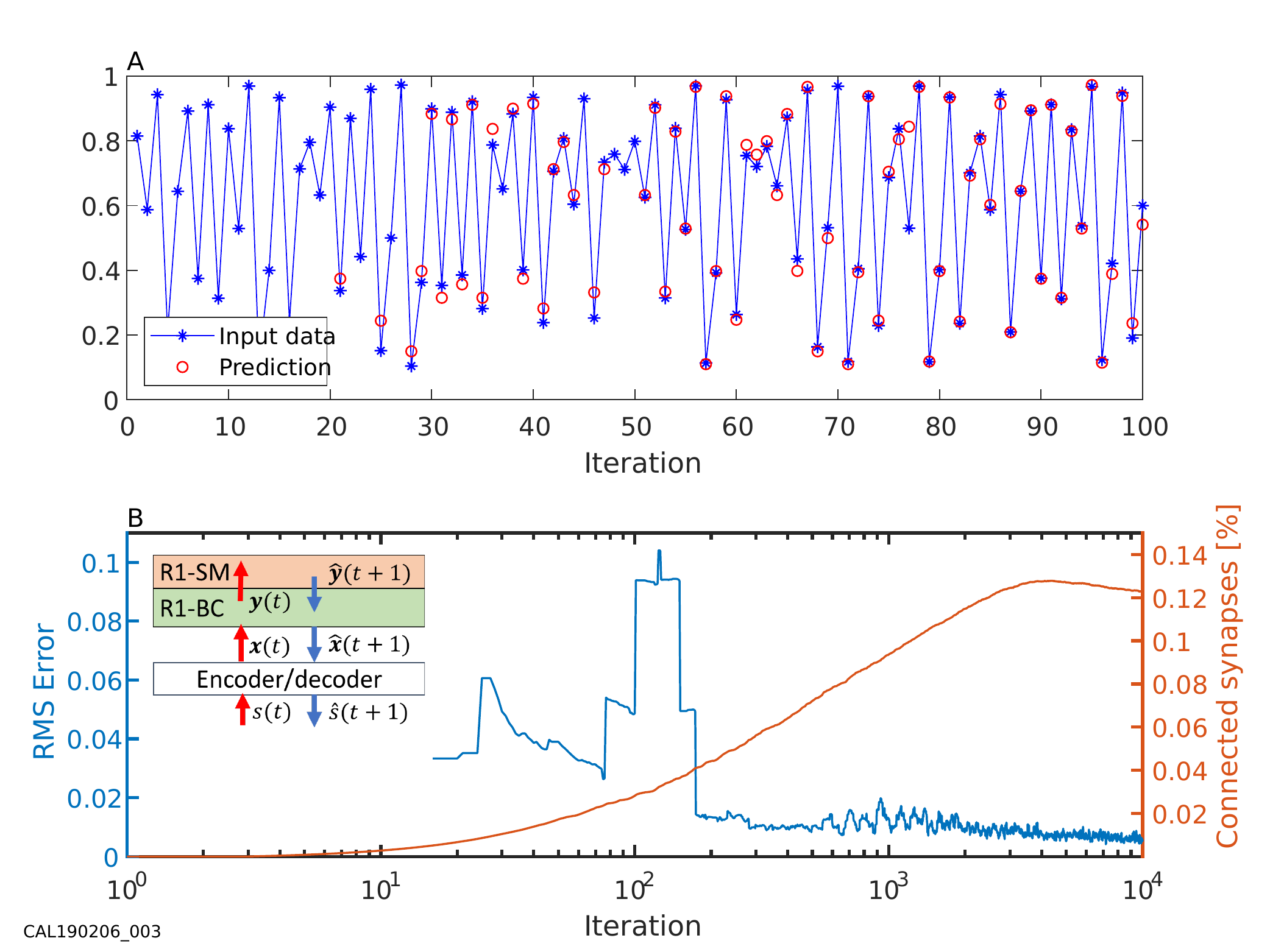}
  \caption{Learning to predict future input of real numbers.  (A) The input data (blue) for the first 100 iterations and the predictions made by the network (red). (B) The root-mean-square error (a running average over 50 iterations) in prediction.  The inset shows the structure of the network (a single region, R1) used for this test, and the flow of data. Scalar input, $s$, encoded as a binary vector, $\mathbi{x}$, enters the binary correlator, BC, which generates output, $\mathbi{y}$, passed to sequence memory, SM.   Based on this and previous input history, SM generates a prediction (indicated by $\hat{ }$, which is reconstructed and decoded. The right scale and red curve show the fraction of connected synapses in SM.}
  \label{fig:popEq}
\end{figure}
The ability to learn sequences and to make predictions based on previous observations is considered a necessary (but far from sufficient) component of intelligent behavior.  In CAL, a synapse array in each region acts as a recurrent neural network to provide sequence memory. It receives input from the binary correlator, which activates one or more neurons in winning mini-columns.  Lateral synaptic connections are modulating\citep{Crick:2003}, i.e. they do not directly activate, but rather predict. via standard multiply and accumulate, which neurons in which mini-columns will be active next. If a column, and predicted neurons in it are indeed then active, the prediction is verified and the responsible synapses are strengthened. The axons of active cells propagate laterally to make predictions of the next input, and also upwards to the next level of the hierarchy.

Sequence memory is conveniently tested using a quasi-chaotic, but deterministic, time-series. A suitable sequence is generated by a simple form of the population equation:
\begin{equation}\label{eq:popEq}
s(t+1)=\beta s(t)[1-s(t)].
\end{equation}

When the parameter 3.57 $\le \beta \le$ 4, $s$ fluctuates chaotically in the range (0, 1); see Fig. \ref{fig:popEq}.  Here we used $\beta=3.89$.  The network that is used to demonstrate learning of this sequence consists of a single region.  Since the input is a single time-series, there is again no need to correlate and synapse updates in the correlator are disabled.
The input-data are encoded binary vectors of size 205, with 5 bits active, resulting in a resolution of 0.005, and fed through the binary correlator (205 $\times$ 1024) which only serves to randomize the $k$=32 active mini-columns of sequence memory which has 8 cells per column and 4 segments per cell.
The results are shown in Fig. \ref{fig:popEq}.  The first valid prediction is made on the 16th iteration, with an error of 0.03.  By the 75th iteration a prediction is being made for every input, but the RMS error has risen as previously unseen values entered the network. Soon the error falls again and after about 1000 iterations settles below 0.01 which is about 5x the expected digitization error.  The prediction error can be reduced closer to that lower limit by increasing the size of the network ($N_{col}, N_{cell}, N_{seg}$), particularly the number of mini-columns, $N_{col}$.

\subsection*{Temporal pooling.}
\begin{figure}[!b]
  \centering
  \includegraphics[width=10cm]{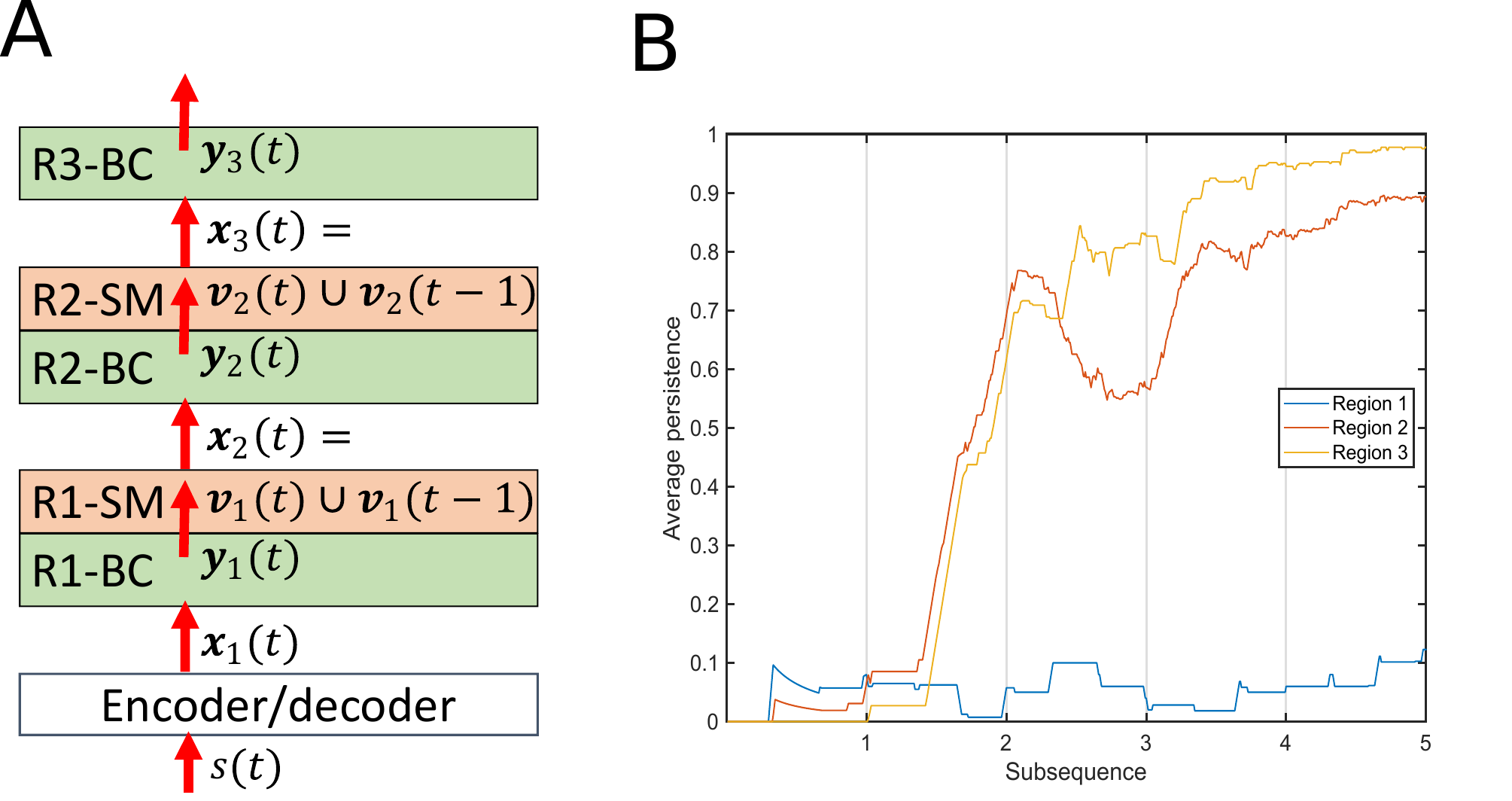}
  \caption{Stability of representation as data propagates up the 3-level hierarchy (A). Notation ($s, \mathbi{x}, \mathbi{y}$) is the same as in Fig. \ref{fig:popEq}, adding the temporal pooling of verified cells, $\mathbi{v}$, and subscripts to indicate hierarchical level. (B) Shows development of ``persistence'', in upper levels of the hierarchy, plotted as a running average over 50 characters.  Each subsequence is a 130 character string, three sentences taken in random order.}
  \label{fig:Persist}
\end{figure}
The bottom level of the hierarchy receives sequences of binary encoded scalar values. Higher regions receive binary vectors indicating the verified cells in the mini-columns of region(s) in the level below. In addition to spatial pooling (concatenation of vectors) these upper regions perform temporal pooling, that is several \textit{successive} vector inputs are combined by union (logical OR).  The output of the correlator then reflects the correlations between time-steps in the sequence(s) learned by the region(s) below. Correlator output enters sequence memory which therefore is learning and representing ``sequences of sequences.'' The verified cell activity passes, in turn, to the next level, where the process is repeated; and so on up the hierarchy.

In mammals, neural activity becomes increasingly stable as data propagates upwards through the levels of sensory cortex to the inferior temporal cortex \citep{DiCarlo:2012}.  We test for this behavior in a network of three regions, arranged in a simple stack: encoded data enters region R1 on level 1; R1 feeds forward to R2, which in turn feeds R3. The input data for this test are three English sentences.  Each sentence is about 45 characters in length, and each set of three is arranged in random order.  To assess stability of representation, $\mathbi{y}$, formed by the correlator, the metric is the fraction of output neurons which remain active in successive time-steps using Jaccard similarity. It is labeled ``persistence'' to emphasize its relationship to time, i.e. persistence is
\begin{equation}\label{eq:Persist}
J \left( \mathbi{y}(t), \mathbi{y}(t-1) \right) = \frac{\mathbi{y}(t-1) \cap \mathbi{y}(t)}{\mathbi{y}(t-1) \cup \mathbi{y}(t)}
\end{equation}

Fig. \ref{fig:Persist} shows the results. At the lowest level, R1, the input character changes on nearly every step, and the persistence merely reflects the averaging time. The persistence increases rapidly as the data propagates to higher levels. After only 5 sets of 3 sentences, the persistence in R3 is 0.98.

The demonstrated formation of increasingly invariant representations in higher regions is in satisfying accord with observations in neuroscience \citep{Quiroga:2005, Cooke:2015, Tsunoda:2001}.  Indeed, according to DiCarlo\citep{DiCarlo:2012} ``core recognition \textit{requires} invariance.'' (Emphasis added.) In CAL, invariance is an emergent consequence of correlating temporally pooled vectors of verified predictions.

\subsection*{Generation of invariant representations.}
In this section we describe an experiment in which a larger CAL network generates invariant representations of various video sequences.  This experiment also demonstrates the scaling of the network, here to 15 regions on 4 levels. It was run on custom, highly-parallel hardware (IBM Neural Computer, INC) that is described in detail elsewhere \citep{Cox:2019}.

\begin{figure}[h]
  \centering
  \includegraphics[width=\linewidth]{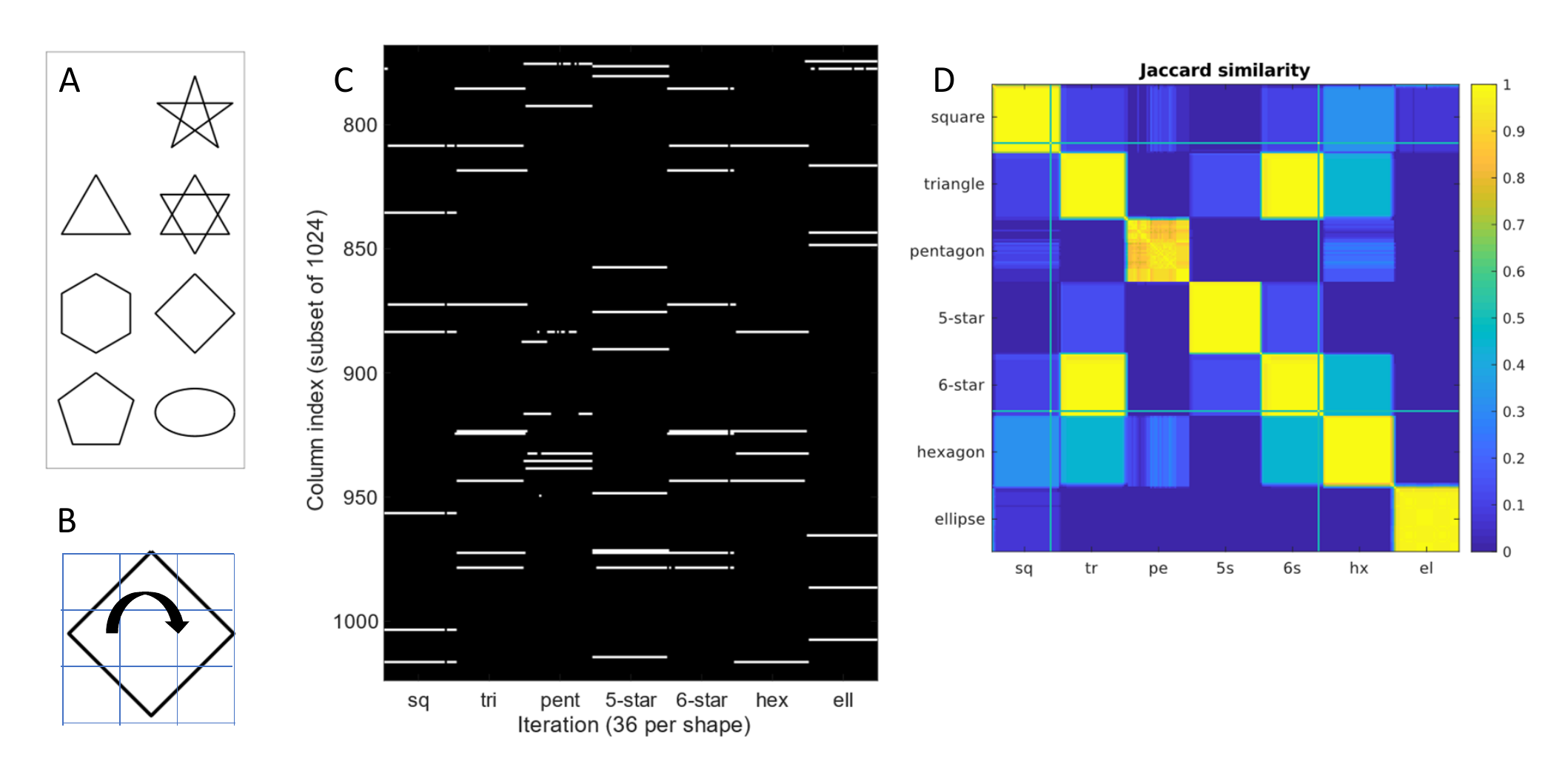}
  \caption{Generation of stable representations.  (A) Input data are binary images of seven different rotating shapes. The network is sketched in Fig. \ref{fig:Scales_4}A (B) Each frame is divided into 9 separate fields and enter 9 regions on the lowest level if the hierarchy.  For each shape, a sequence is 36 successive frames.  (C)  After about 200 epochs, each 7 $\times$ 36 iterations, the output of the level 4 correlator is essentially constant for as each shape enters level 1.  Each pixel represents the activity of a single column (y-axis) during one time-step (x-axis).}
  \label{fig:RotatingShapes}
\end{figure}

The input data are frame-by-frame binary images of seven different rotating shapes Fig. \ref{fig:RotatingShapes}A.  One such frame, a square rotating about its center, is shown in Fig. \ref{fig:RotatingShapes}B.  The other shapes are hexagon, pentagon, triangle, 6-pointed star, 5-pointed star, all of which rotate about their center, and a circle which rotates about a horizontal in-plane axis and so is generally seen as elliptical except when it is face-on or edge-on. Each frame is divided into 9 non-overlapping receptive fields, Fig. \ref{fig:RotatingShapes}B and the binary pixel arrays are converted, column-major, to a vector.  No further encoding is required.

The network architecture is shown, schematically, in Fig. \ref{fig:Scales_4}A. (Some feed-forward and feedback paths are omitted for clarity.)  The vector of each receptive field is fed directly into one of 9 first level regions.  Each of the 7 shapes is shown for a full rotation, 36 frames at 10 degree steps, so an epoch is $7 \times 36 = 252$ iterations. The four regions on level-2 receive input from four adjacent regions on level-1.  Thus the regions of level-2 correlate, both spatially and temporally (over 3 consecutive time-steps), the verified predictions of four adjacent receptive fields.  A single level-3 region processes the pooled output of the four regions on level-2, and one region on level-4 is the temporal correlator for verified cells of level-3.

Correlator output at the top level, during the last of 200 epochs is shown in Fig. \ref{fig:RotatingShapes}C. The abscissa is labeled with the shape being shown, adjusting for the propagation delay up the 4 levels.  The level-4 representation is very stable for each shape, and generally differs from (i.e., is nearly orthogonal to) those of other shapes.  The Jaccard similarity between output vector pairs, Fig. \ref{fig:RotatingShapes}D, shows the stability, similarities and differences of and between representations. Blocks on the diagonal reveal that all representations, with the exception of the pentagon, are virtually invariant while each shape enters the network.  Off-diagonal blocks show that the closest match is between equilateral triangle and 6-point star, which is the overlay of two such triangles. The pentagon shows intermittent similarity to square and hexagon, i.e. the shapes with respectively one fewer and one more edge. The ellipse is completely orthogonal to all but the square, both sometimes presenting horizontal lines.

Thus this 4-level network does more than just auto-encode the rotating shapes:  it generates its own ``labels'' in the form of distinct invariant vector representation of each sequence, where similarities in shape result in common active bits.  This is analogous to the recognition of objects, people, animals and events by infants and non-verbal species.  The integrity of each entity is established in some representative form along with characteristic features, even though the individual may not have the vocabulary to express the concept \citep{Harnad:1990, Barth:2012}.

\section*{Discussion.}
The network behavior of CAL relies on Hebb's original postulate\citep{Hebb:1949}, modified to include anti-Hebbian weakening. The rules  remain entirely local, involving only one axon, one dendrite and the synapse that connects them.  This avoids calculating gradients in neural excitation for back-propagation of errors and the need for large datasets and tens of thousands of iterations.  Learning occurs rapidly, in about 100 iterations, in real time, based on each observation in the context of what preceded it.  As a result, time-averaged correlations of the input are learned in on sub-network (the binary correlator) of each region, as proved by Oja\citep{Oja:1982}, and sequences are learned in a second, laterally connected, recurrent sub-network (sequence memory). The third sub-network in each region of CAL, the apical array receiving feedback, was active in the demonstration of stable representations, and hints that feedback can accelerate such learning.  Preliminary experiments show that feedback helps to mitigate noisy and missing data in the input. This is the subject of ongoing work.

The results described in the Results section illustrate the necessity of simulating complex neurons.  In particular, by distinguishing dendrites that modulate the response of neurons, as opposed to causing immediate firing\citep{Crick:2003}, the system rapidly learns of recurring sequences via update of synapse weight responsible for verified and false predictions.  Modulation is also used in feedback.

The large scale structure of CAL mimics the well established connectionist view of the brain\citep{Felleman:1991, Hubel:1959, Yildirim:2019}. Within this corticothalamic network\citep{Nakajima:2017}, neural processing occurs by propagation of data through the network via axons activated by firing neurons. This activity causes changes in network topology, by creating of new synapses, as well as in the weights of existing synapses.  Synaptic plasticity, in both existence and weight of connections, is critical to the efficient operation of this process.  The neural firing patterns are data representations and they create memories that are held by synapses.  In turn, these memories are retrieved, and propagated by neural activity\citep{Albo:2018}.

Several aspects of biological homeostasis and self-regulation have been incorporated into CAL.  For example, when new synapses are needed, due to activity in previously inactive axons, they are created on dendrites with the fewest existing connections. This ensures optimal use of network resources, reflecting the biological cost in terms of energy and biochemical resources of overworked cells.  Similarly, by making anti-Hebbian weight decreases proportional to Hebbian increases, according to the numbers of synapses increased and decreased, the network is stabilized against oscillations and runaway.  A simple application of meta-plasticity\citep{Abraham:2008}, namely reducing anti-Hebbian decrements in the most permanent synapses, virtually eliminates ``catastrophic forgetting''.\citep{French:1999}

Neural activity (i.e., $k$ in $k$-winners-take-all) is set by consideration of the statistics which govern accidental matching of binary vectors. The hypergeometric distribution yields a monotonically decreasing probability of more than a single bit randomly matching if $k <= \sqrt N$, where $N$ is the length of the vector. Therefore, to achieve good capacity ($N$-choose-$k$), while minimizing accidental matching, we set $k = \sqrt N_{col}$.

The homeostatic and statistically guided features of the algorithms have the additional benefit of reducing the number of meta-parameters that must be set and tuned by the user.  For each region, only four meta-parameters do not take default values: the number of mini-columns, the number of neurons in each mini-column, the number of dendritic segments for each neuron, and the learning rate. Of course, the code allows the user to override the defaults.

The behavior of the network does not arise from deliberate design: rather it emerges as a result of applying these established neurological principles. Correlation arises from the application of Hebbian learning rules to concatenated representations in the static case and to union of consecutive representations in the temporal case\citep{Oja:1982}.  Generation of invariant representations in the upper levels of the hierarchy is the emergent consequence.

The reader may wonder how humans can ever interpret the network's representations which are apparently random vectors (e.g. Fig. 4C). Remembering that animals and infants understand concepts without the vocabulary to represent them, one strategy would be to replace ``supervision'' using labeled and curated datasets, with ``language instruction'' providing the network with (binary vector) representations of words and phases that it then correlates with its internally generated vector symbol of the concept.  Preliminary studies in this direction are promising: we have shown that if the correlator is trained with the concatenation of vectors $\mathbi{a}$ and $\mathbi{b}$, then subsequently the input is only $\mathbi{a}$, the reconstruction of its output includes $\mathbi{b}$. The network has learned to associate $\mathbi{a}$ with $\mathbi{b}$.

We close by returning to the concept of modulation.  In accord with neuroscience, the idea can be generalized beyond the electrical modulation that reduces the firing threshold of the neuron.  Threshold is just one of many network meta-parameters; its reduction is depolarization induced by ionic neurotransmitters.  Other, non-ionic, neurotransmitters are responsible for a wide variety of responses: for example several modulate the level of neural activity (the $k$ winners), either raising or lowering it \citep{Purves:2018} Ch. 6. Dopamine modulates the magnitude of synapse weight changes\citep{Reynolds:2002}, i.e., the learning rate parameter.  CAL provides a framework for exploration of these ideas, hopefully benefitting the fields of both artificial intelligence and neuroscience.

\section*{Methods}
\subsection*{Architecture.}
\begin{figure}[h]
  \centering
  \includegraphics[width=\linewidth]{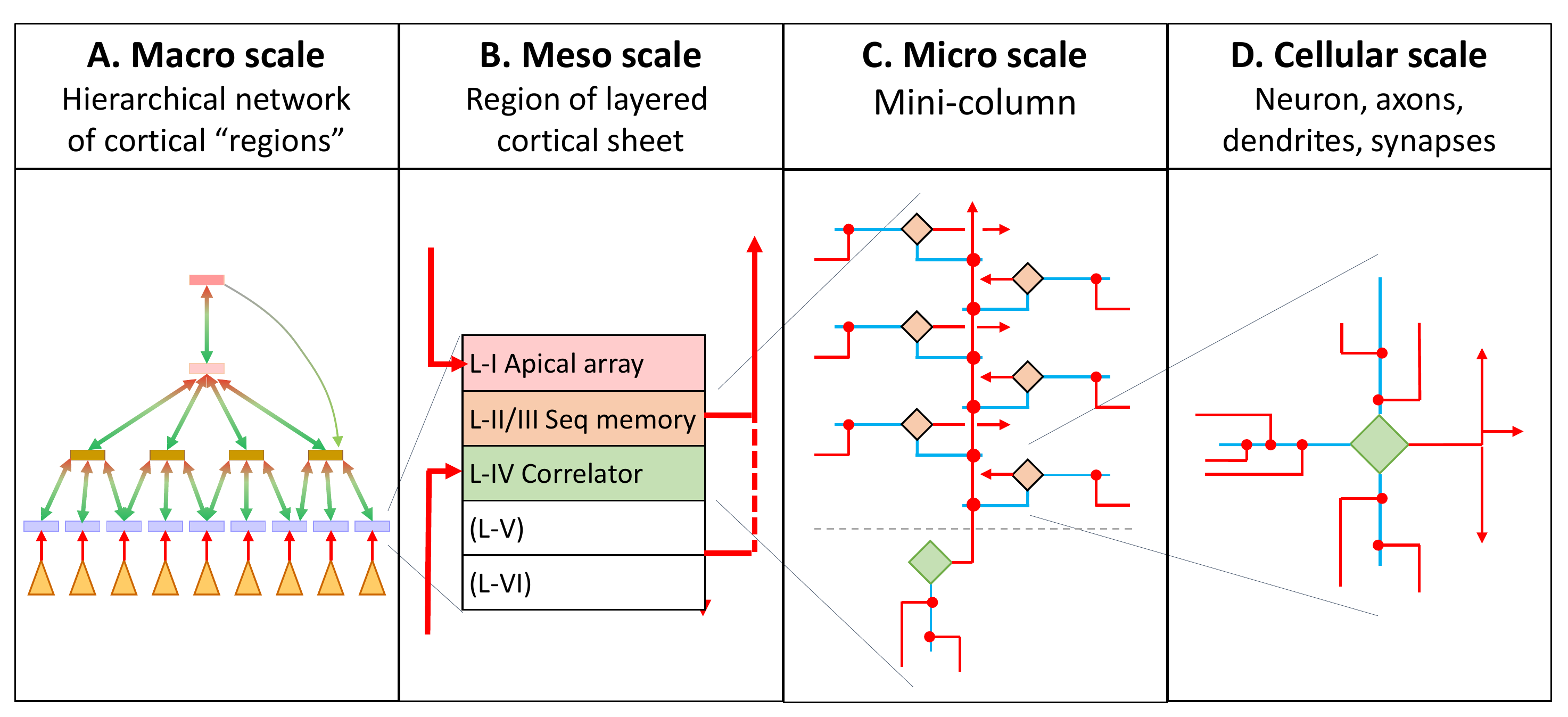}
  \caption{Four length scales of cortical organization. (A) Macro-scale $\sim$10cm. Cortical regions are interconnected in a hierarchy.  Axons feed-forward from lower regions to regions in the next level. The bottom level receives input data-streams (''sensory channels'') encoded as binary vectors. Feedback propagates from higher regions to any region below. (B) Meso-scale $\sim$10mm. Each region is a piece of the cortical sheet, consisting of six cortical layers. Currently, layers L-I to L-IV are simulated. The dendrites of layer L-IV neurons connect to feed-forward axons propagating from lower levels, and act as the correlator. L-II-III has many lateral modulating (predicting) connections and acts a sequence memory. Layer L-I, has few neurons, but contains the apical dendritic segments of neurons in other layers, that connect to axons carrying feedback from higher levels of the hierarchy. (C) Micro-scale, $\sim100\mu$. Neurons in L-IV and L-II/III are organized into mini-columns, defined as the set of L-II/III neurons which are driven by the same L-IV neuron.  Multiple mini-columns are packed side-by-side to form the cortical sheet. (D) Cellular scale, $\sim5\mu$. Each a neuron has one or more dendrites (blue) and a branching axon (red) that carries outgoing information about its state of activation. Dendrites are classified as basal, receiving input via afferent axons from lower in the hierarchy, lateral from the same level, or apical from above. Synapses connecting axons and dendrites are red dots.}
  \label{fig:Scales_4}
\end{figure}

The structural organization of CAL is illustrated in Fig. \ref{fig:Scales_4}.  Four different length scales, based on the anatomy of the mammalian cortex, are defined.  It is helpful to describe the structure starting at the largest scale, and ``dissecting'' down to the cellular level.  After the structural organization has been described, the function and processes will be detailed.

At the largest scale, the cortical sheet (grey matter) forms the entire outer covering of the brain, and is located just inside the skull.  Thanks to a range of studies starting with tracing pathological loss of function in the 19th century \citep{Broca:1861} through modern functional-MRI \citep{Guy:1999, Kahnt:2017, Haile:2019} it has become clear that different areas of the cortical sheet are involved in different neural processes, from interpreting sensory data at the most primitive level, to logical reasoning and language at the highest.  Moreover, data tend to flow sequentially between these areas in a hierarchical fashion \citep{Felleman:1991, Riesenhuber:1999}, especially from the sensory cortex to the mid-brain.  In CAL, we use the term ``region'' to identify the nodes of this hierarchy, Fig. \ref{fig:Scales_4}A, mainly to avoid identification and confusion with Brodmann ``areas'' \citep{Brodmann:1905}, since there is no attempt, yet, to simulate a one-to-one mapping of region with function.

Each region contains many neurons, whose axons run throughout the volume of the brain (the white matter) carrying data from one region to another, often via the thalamus.  Although the cortex is a single (crumpled) sheet, we emphasize the long range connectivity by separating the nodes/regions and representing axon bundles (fibers) as arrows.  Sensory data enters at the lowest level where it is processed and fed forward level by level.  In turn, output of upper levels, may also be fed back to regions in any lower level.

Within each region, Fig. \ref{fig:Scales_4}B the cortical sheet consists of multiple (up to 6) layers that are differentiated by the density and type of their neurons.
Currently, CAL simulates layers the processes that are believed to occur in L-I, L-II/III and L-IV.  (We use Roman numerals to distinguish the anatomical \textit{layers} of the cortex from the logical \textit{levels} of the hierarchy.)

Feed-forward input, via axons propagating from lower in the hierarchy, enters layer L-IV. The activity of L-IV neurons propagates via axons into L-II/III of the same region, as well as to lower regions of the hierarchy.  In turn, axons in L-II/III propagate both laterally within the regions as well as upwards in the hierarchy, either directly or via the thalamus. Feedback activity is carried by axons that enter layer L-I, which has few neurons (none in our model) but many synapses on dendrites belonging to neurons in other layers of the region.

At the next scale Fig. \ref{fig:Scales_4}C, we find the organization of neurons, notably their connection into mini-columns that run perpendicular the cortical sheet.  Each mini-column consists of a single ``master'' neuron in cortical layer L-IV and number of neurons in LII/III that (for simplicity) connect to the L-IV cell through a single permanent synapse.  A mini-column is then identified as the set of L-II/III neurons which share the same L-IV axon. LII/III neurons connect with each other via axons that run laterally within the region.  Several hundred mini-columns packed side-by-side in the cortical sheet, form a region.

At the cellular scale, Fig. \ref{fig:Scales_4}D, we finally see the structure of a neuron.  The drawing emphasizes the role of dendrites in neural function. Each cortical neuron receives input from up to three distinct sources via dendrites that are labeled accordingly: feed-forward data, either from lower regions in the hierarchy within the region (e.g. L-IV to L-II/III), is received via basal dendrites; axons propagating from neurons within layer L-II/III of the same region connect to lateral dendrites; and feedback from upper levels of the hierarchy is received by apical dendrites that typically extend into layer L-I. (Due to their long reach, these are also called distal dendrites.)  Basal dendrites are ``driving,'' that is their excitation may cause immediate firing of the neuron depending on its state which is dictated by ``modulating'' lateral and/or apical dendrites\citep{Crick:2003}.  The early concept of a point neuron\citep{MCP:1943} that integrates and fires, is replaced by multiple dendrites that either ``integrate and drive'' or ``integrate and modulate.''

In each short time-interval, dendritic activity on each neuron either cause it to fire, or not, in an ``all or nothing'' (binary) event.  The activity of the neurons is then carried by its axons to rest of the network.  The complexity of the neurons then permits the neural processes that we now describe.  Three arrays of synapses are currently simulated in CAL, corresponding to the three colored layers in Fig. \ref{fig:Scales_4}B. CAL is a ``network-of-networks.''

The basal synapses of the L-IV neurons form the simplest array.  Each afferent axon from regions lower in the hierarchy or, in the bottom level, from sensory inputs, \textit{potentially} connects to the (single) basal dendrite of any neuron.  Every synapse has a weight indicating how strongly it is connected, ranging from zero (not connected) to a maximum of unity.  If a dendrite is sufficiently well connected to a sufficient number of active axons (``sufficient'' is defined in the algorithms section) then its neuron fires.  Synapse weights are then updated according to Hebbian rules: if both the axon and dendrite were active, the synapse is strengthened; if only the axon, or only the dendrites was active, it is weakened (anti-Hebb); if neither was active, there is no change.
Collectively, the synapses on the L-IV dendrites form a correlator as a result of the Hebbian and anti-Hebbian updates as shown by Oja \citep{Oja:1982}.

Axons of neurons in layer LII/III connect laterally to dendrites in the same layer.  These lateral connections do not cause immediate firing, but rather ``predict'' the neurons that will fire next due to feed-forward input.  This is modulation, in the form of lowering the firing threshold.  If the prediction is confirmed, the synapses that contributed are strengthened; if a neuron was predicted, but did not fire, the relevant synapses are weakened.  By this means, LII/III learns recurrent sequences in the input data \citep{Hawkins:2017}.

Typically, by inhibition, only a single neuron in each mini-column is predicted.  Each prediction was generated by previously active cells in previously active columns.  Thus the particular L-II/III neuron that fires, when the L-IV neuron activates it, indicates the context in which the prediction was made.  Then, if that neuron contributes to the next verified prediction, the context is extended to the next time-step.  To state it another way, after a sequence has been learned, when the mini-columns representing 'B' are active, the cells that fire are those representing 'B' in the context of 'A.'  Next, when the input is 'C', the active cells in the 'C' mini-columns are those representing context 'A-B.'

The array of synapses in layer L-I potentially connect to the the apical dendrites of neurons in L-II/III.  Their axons propagate from regions in higher levels of the hierarchy, i.e. they receive feedback.  Like lateral dendrites, apical dendrites are modulating, and the feedback activity is likewise predictive.  However, because of propagation delays, the predictions are made on the basis of input at earlier time-steps.

Currently the network designer must decide how to make region-to-region connections, upward and downward in the hierarchy.  This fixed routings play the role white-matter and the thalamus.\citep{Purves:2018} Box A-17.  In the future, it may be necessary to simulate a biological thalamus with neurons and plastic synapses that learn the routing connections.

\subsection*{Algorithms.}
The code for CAL has been written in both Matlab\textsuperscript{TM} and c++, using objects which correspond to the major features illustrated in Fig. \ref{fig:Scales_4}. The primary objects correspond to the three synapse arrays in each region - Fig. \ref{fig:Scales_4}B. Synapse weights are represented as a sparse matrix, indexed by axon and dendrite.  The binary correlation (i.e. the basal array of L-IV neurons) receives feed-forward input as (possibly concatenated) binary vectors.  The activity L-IV neurons drives the L-II/III neurons of their respective mini-columns.  However, because of modulation by lateral dendrites (the array of sequence memory), only a few cells in L-II/III fire.  A single axon from each L-II/III neuron potentially interconnects laterally to any lateral dendritic segment of any other neuron.  The apical array, in layer L-I, has axons that originate in upper levels and dendrites that modulate the response of neurons in layers L-II/III and L-IV.

\subsection*{Encoding and decoding.}
All regions of CAL operate on binary vectors, whereas much input data is a time-series of scalars. Therefore at every iteration and for for each input channel values, $s(t)$, are encoded using a ``slide-bar encoder'' \citep{Hawkins:2011}, where a number of adjacent bits (the bar) of a binary vector are turned on, and the position of the bar within the vector (the slide) encodes a scalar value. Having more than one active bit provides redundancy and ensures more robust decoding.

For every scalar channel, the user must decide on the length ($N$) of the binary vector, and the number of bits ($1 \leq k \ll N$) that are active, based on the range of the data ($R = s_{max} - s_{min}$), and the desired resolution ($r$).  For real numbers, this should be comparable to the noise level for real numbers. For integers, $r = 1$.  The number of encoded values is then $R/r + 1$.  Each encoding encompasses a range of values, a ``bin'', $(-r/2, +r/2)$ relative to the digitized value, and input value, $s$, falls into bin number $b = \textnormal{Round}((s - s_{min})/r)$.

Real numbers that are close in value should have similar encoding, and therefore adjacent bins may overlap, meaning that the displacement, $d$, between neighboring bins is less than the number of active bits, $k$. On the other hand, integers should be entirely distinct from each other and not overlapping, (for example the ASCII encoding of `a' and `b', 97, and 98 respectively), so the displacement, $d = k$. These parameters set the vector size $N = dR/r + k$, with only the bits ranging from $bd + 1$ to $bd+k$ (index origin 1) turned on.

It is convenient for decoding to construct a look-up-table (LUT) in the form of a sparse binary matrix, where the columns are indexed by bin-number and contain the corresponding vector encoding.  Then, the first step in decoding is to find, by matrix multiplication, the column (i.e. bin, $\hat{b}$) which best matches the vector to be decoded. The corresponding (real or integer) value is then given by $\hat{s} = \hat{b} \times r + s_{min})$.

In addition to integer and real numbers, CAL accepts directly, without encoding, sequences of binary 2D images.  Each frame of the sequence is a matrix which is simply unwrapped in column-major order.  Typically, line-drawings are already suitably sparse; other images can be preprocessed using segmentation, edge-detection, etc.  CAL is agnostic to the type of the input data.  Once encoded into sparse binary vectors, all data are processed in the same way.  Multiple input channels to a single region are combined (spatially pooled) simply by concatenating their binary encoding.

\subsection*{Synaptic weight and permanence.}
In any array, each synapse is identified by the indices of the axon, $i$, and dendrite, $j$, that it potentially connects.  It has two (closely related) properties:  weight, $w_{ij}$, determines how efficiently it passes electrical stimulus from axon to dendrite; permanence, $p_{ij}$, is a measure of how ``well established'' it is, and corresponds roughly to physical size which is observed in neuroscience to increase as the synapse is repetitively strengthened by Hebbian updates.  The weight is simply a quantized, reduced precision, approximation of permanence, and may be thought of as the number of ion-channels in the membrane of synaptic cleft. CAL has been tested with weight precision of 1-bit (i.e. binary), 2, 3, 4, 8 and full floating-point.

\subsection*{Activation with inhibition.}
The input and output of a synapse array with $m$ axons and $n$ dendrites is represented by binary column vectors
$\boldsymbol{x} \in \mathbb{B}^{m \times 1}$ and $\boldsymbol{y} \in \mathbb{B}^{n \times 1}$, respectively.
The weight matrix, $\boldsymbol{W} \in \mathbb{R}^{m \times n}$. Here and in the following, bold, italic font is used to indicate vectors and matrices, with the latter capitalized.  Scalars are italic not bold.

Then the multiply, accumulate and threshold, with \textit{k}-winners-take-all inhibition can be written in matrix notation as
\begin{equation}
\mathbi{y} = (\mathbi{x} ^{\mathsf{T}} \mathbi{W}) ^{\mathsf{T}} \geq \tau; \; \tau \; \textnormal{such that} \; |\mathbi{y}| = k.
\label{eq:kWTA}
\end{equation}
where $|.|$ indicates cardinality (aka ``pop-count''), and the Boolean result of $\geq$ is applied element-wise to the vector-matrix product.  $\tau$ is the (automatically adjusting) threshold. For computational efficiency, transpose ($.^\mathsf{T}$) is applied to vectors, not matrices.

\subsection*{Synapse updates.}
Given the input activity, $\mathbi{x}$, and output activity, $\mathbi{y}$, of a synapse array, permanence updates are calculated according to
\begin{equation}
\delta\mathbi{P} = \delta_{AA} \mathbi{xy}^{\mathsf{T}} - \delta_{AI}\mathbi{xe}_{n}^{\mathsf{T}} -
\delta_{IA}\mathbi{e}_{m} \mathbi{y}^{\mathsf{T}}
\label{eq:update}
\end{equation}
where the subscripts on coefficients $\delta$ indicate whether both axon and dendrite were active (Hebbian, $AA$), only the axon (anti-Hebbian, $AI$) or only the dendrite (anti-Hebbian, $IA$), and where $\mathbi{e}_{m},\mathbi{e}_{n}$ are column vectors of all 1s, length $m$ and $n$, respectively.

\subsection*{Binary correlator.}
The binary correlator of each region is a simple array of synapses.  The axons carry input data from other regions lower in the hierarchy, or in the case of the first level, ``sensory'' data from one or more encoders.  The (here, unsegmented) driving dendrites belong to the neurons of cortical layer L-IV. There is a \textit{potential} synapse at the intersection of each axon and each dendrite, but very few of them are actually connected.

Since each neuron has just a single dendrite, equation (\ref{eq:kWTA}) applies directly to the activation of the neurons, $\mathbi{W}$ refers to the weights in the correlator, whose input is $\mathbi{x}$, output $\mathbi{y}$.  Similarly, the updates, equation (\ref{eq:update}), are applied to the permanence matrix, $\mathbi{P}$, of the correlator.

\subsection*{Reconstruction.}
The set of mini-columns predicted in sequence memory must be reconstructed and decoded to scalar values of the same data-type and range as the input channel.  Reconstruction is simply the reverse of the correlation process.  The predicted vector, $\hat{\mathbi{y}}$, is multiplied by the transpose of the correlator weight matrix.  The predicted binary input, $\hat{\mathbi{x}}$, is found by taking the $k_{in}$, the number of active bits in the vector input, largest elements of the result,
\begin{equation}
\hat{\mathbi{x}} = (\mathbi{W} \hat{\mathbi{y}}) \geq \tau; \; \tau \; \textnormal{such that} \; |\hat{\mathbi{x}}| = k_{in}
\label{eq:recon}
\end{equation}
analogously to equation (\ref{eq:kWTA}).  $\hat{\mathbi{x}}$ is then decoded to give the predicted scalar value, $\hat{\mathbi{s}}$

\subsection*{Sequence memory.}
In each region, sequence memory consists of $N_{col}$ mini-columns each with $N_{cell}$ neurons per column for a total of $N_{cell}N_{col}$ neurons.  Each neuron has $N_{seg}$ modulating, lateral dendrite segments which accumulate the activity of connected axons.  If a segment is sufficiently excited, above a self-adjusting threshold, its neuron is predicted to be active at the next time-step.  A single axon propagates from each neurons to form a sparse, laterally connected recurrent network.  The dimensions of the matrix representing \textit{potentially} connected synapses is therefore $N_{cell}N_{col} \times N_{seg}N_{cell}N_{col}$.

At time $t$, the inputs to sequence memory are the output of the region's correlator, $\mathbi{y}(t)$, and the vector representing the neurons predicted in the \textit{previous} time-step $\mathbi{z}(t-1)$.  The outputs are the vector of verified neurons $\mathbi{v}(t)$, and of predicted neurons $\mathbi{z}(t)$.  For synapse updates, the algorithm also requires that the system retain a vector representing the excitation of the dendrites, $\mathbi{d}(t)$ in order to identify those responsible for predictions.

The algorithm proceeds in four steps: first the verified neurons are identified, that is the neurons in mini-columns that are now active and were previously predicted.
\begin{equation}
\mathbi{v}(t) = \textnormal{reshape}(\mathbi{I}_{N_{cell \times 1}} \mathbi{y}(t)^\mathsf{T}, \: N_{cell}N_{col}, \: 1) \; \& \; \mathbi{z}(t-1)
\label{eq:verify}
\end{equation}
where $\mathbi{I}_{N_{cell \times 1}}$ is a column vector, length $N_{cell}$, of all 1's, and the reshape function converts its matrix argument to a column vector.

There are two categories: the neurons that are verified, $\mathbi{v}(t)$; and all of the neurons in unpredicted mini-columns. The second category (referred to a ``bursting'' behavior by the Numenta group \citep{Hawkins:2011}) enables the system to make exploratory predictions that increase the probability of finding a subsequence when an active mini-column has not been predicted.

For category (ii), to find the unpredicted mini-columns, reshape the vector of predicted neurons into a matrix $\mathbi{Z}$, with elements $\zeta_{i,j}$
\begin{equation}
\mathbi{Z} = \textnormal{reshape}(\mathbi{z}, \: N_{cell}, \: N_{col})
\label{eq:predcolmat}
\end{equation}
Then the predicted columns are
\begin{equation}
\mathbi{c} = \left( ( \sum_i \zeta_{i,j} ) > 0 \right)^\mathsf{T}
\label{eq:predcol}
\end{equation}
and those unpredicted are
\begin{equation}
\mathbi{u} = \mathbi{y} \: \& \: \neg\mathbi{c}
\label{eq:unpredcol}
\end{equation}

The bursting cells are found by replication of $\mathbi{u}$ and unwrapping in column-major order. Then neuron activity is
\begin{equation}
\mathbi{b} = \textnormal{reshape}(\mathbi{I}_{1 \times N_{cell}} \mathbi{u}^{\mathsf{T}}, \: N_{cell}N_{col}, \: 1)
\label{eq:burst}
\end{equation}

Then neuron activity is
\begin{equation}
\mathbi{a} = \mathbi{v} \; | \; \mathbi{b}
\label{eq:smactive}
\end{equation}

The axons of active cells excite the lateral dendrite segments, according to the weights of connected synapses, i.e.
\begin{equation}
\mathbi{d} = \mathbi{a}^{\mathsf{T}} \mathbi{W}_{SM}
\label{eq:segment}
\end{equation}
Note that $\mathbi{d}$ is a column vector of real numbers, not binary. Since $k$ columns will be activated at the next step, a vector representing $k$ active segments is found by the winners-take-all algorithm,
\begin{equation}
\mathbi{s} = \mathbi{d} \geq \tau; \; \tau \; \textnormal{such that} \; |\mathbi{s}| = k.
\label{eq:predictseg}
\end{equation}

The segments of bursting cells which came closest to being predicted (largest element of $\mathbi{d}$) are found and added to the list of those that actually made the predictions. If all $d_i$ are zero, one is chosen at random.  Since there are $D=N_{seg}N_{cell}$ dendritic segments associated with each column, the  pseudo-code (index-origin zero) is:
\vspace{6pt}
\hrule
\vspace{3pt}
for $j$ in range($N_{col}$)
\\  \indent if $u_j$
\\  \indent \indent  if $\sum d_{jD:(j+1)S} == 0$
\\  \indent \indent  \indent  $i$ = randi($D$)
\\  \indent \indent  else
\\  \indent \indent  \indent  $i$ = argmax($d_{jD:(j+1)D}$)
\\  \indent  $s_{jD} + i$ = true
\vspace{3pt}
\hrule
\vspace{6pt}
Any neuron that has an active dendrite is predicted, therefore, we sum over segments of each neuron:
\begin{equation}
\mathbi{z} = ( \sum_i \textnormal{reshape}(\mathbi{s}, \: N_{seg}, \: N_{cell})) > 0.
\label{eq:predictcell}
\end{equation}

Synapse updates are made on the basis of axons which were active and contributed to verified predictions, or not.  Thus, in equation \ref{eq:update}
\begin{equation}
\mathbi{x} = \mathbi{a}(t-1)
\end{equation}
\begin{equation}
\mathbi{y} = \mathbi{s}(t-1)
\end{equation}

\subsection*{Temporal pooling.}
The vector output, $\mathbi{v}(t)$, of sequence memory represents the verified neurons, those predicted at the previous time-step and now active and is fed-forward to become the input to regions on the next highest level. Temporal pooling is simply the union of two or more consecutive vectors, thus the input to a region on level $L$ is

\begin{equation}
\mathbi{x}_L(t) = \mathbi{v}_{L-1}(t) \: \cup \mathbi{v}_{L-1}(t-1) \: \cup \mathbi{v}_{L-1}(t-2) \: \ldots
\label{eq:temporalPool}
\end{equation}

where $\mathbi{v}_{L-1}$ is the concatenation of all output vectors $\mathbi{v}(t)$ from level $L-1$.

\subsection*{Feedback.}
The output of each binary correlator is not only the input to sequence memory of the same region, it may also be fed back to regions on any lower level.  Biologically, this occurs via matrix neurons of the thalamus.  Currently, CAL simulates this feedback via axons to layer L-I of the lower region.  These axons potentially connect to apical synapses of neurons in layer L-IV, modulating their response, via lowering of their firing threshold.  Noting that threshold reduction is equivalent to increasing the excitation of the relevant dendrite, it is computationally convenient to introduce a variable, dendritic ``gain'', that implements the modulation causes by feedback $\mathbi{f}$.

\begin{equation}
\mathbi{g} = (\mathbi{f} ^{\mathsf{T}} \mathbi{W}_{Ap} )^{\mathsf{T}}
\label{eq:dendGain}
\end{equation}

Then equation (\ref{eq:kWTA}) is modified according to

\begin{equation}
\mathbi{y} = \left( (\mathbi{x} \odot \mathbi{g}) ^{\mathsf{T}} \mathbi{W}_{BC} \right) ^{\mathsf{T}} \geq \tau; \; \tau \; \textnormal{such that} \; |\mathbi{y}| = k.
\label{eq:modulate}
\end{equation}
where $\odot$ is the Hadamard product, and the subscripts $Ap$ and $BC$ refer to the apical and binary correlator weight matrices, respectively.

In the future apical synapse connections for neurons in layer L-II/III may be added.
\newpage

%% Put the bibliography here
\bibliography{Scott_2019} % bib extension not needed?

\begin{thebibliography}{67}
\providecommand{\natexlab}[1]{#1}
\providecommand{\url}[1]{\texttt{#1}}
\expandafter\ifx\csname urlstyle\endcsname\relax
  \providecommand{\doi}[1]{doi: #1}\else
  \providecommand{\doi}{doi: \begingroup \urlstyle{rm}\Url}\fi

\bibitem[Abraham(2008)]{Abraham:2008}
Wickliffe~C. Abraham.
\newblock Metaplasticity: tuning synapses and networks for plasticity.
\newblock \emph{Nat. Rev. Neurosci.}, 9:\penalty0 387, 2008.

\bibitem[Albo and Graff(2018)]{Albo:2018}
Zimbul Albo and Johannes Graff.
\newblock The mysteries of remote memory.
\newblock \emph{Phil. Trans. Roy. Soc. B: Biol. Sci.}, 373\penalty0 (1742),
  2018.

\bibitem[Azarfar et~al.(2018)Azarfar, Calcini, Huang, Zeldenrust, and
  Celikel]{Azafar:2018}
Alireza Azarfar, Niccoló Calcini, Chao Huang, Fleur Zeldenrust, and Tansu
  Celikel.
\newblock Neural coding: A single neuron’s perspective.
\newblock \emph{Neuroscience \& Biobehavioral Reviews}, 94:\penalty0 238--247,
  2018.

\bibitem[Barth and Poulet(2012)]{Barth:2012}
Alison~L. Barth and James F.~A. Poulet.
\newblock Experimental evidence for sparse firing in the neocortex.
\newblock \emph{Trends in Neurosciences}, 35\penalty0 (6):\penalty0 345--355,
  2012.

\bibitem[Broca(1861)]{Broca:1861}
Pierre~Paul Broca.
\newblock Nouvelle observation d’aphémie produite par une lésion de la
  troisième circonvolution frontale.
\newblock \emph{Bulletins de la Société d’anatomie (Paris), 2e serie},
  6:\penalty0 398--407, 1861.

\bibitem[Brodmann(1905)]{Brodmann:1905}
K.~Brodmann.
\newblock Beitrage zur histologischen localisation der grosshirnrinde. dritte
  mitteilung. die rindenfelder der niederen.
\newblock \emph{Affen. J Psychol Neurol}, 4:\penalty0 177--226, 1905.

\bibitem[Burr et~al.(2017)Burr, Shelby, Sebastian, Kim, Kim, Sidler, Virwani,
  Ishii, Narayanan, Fumarola, Sanches, Boybat, Le~Gallo, Moon, Woo, Hwang, and
  Leblebici]{Burr:2017}
Geoffrey~W. Burr, Robert~M. Shelby, Abu Sebastian, Sangbum Kim, Seyoung Kim,
  Severin Sidler, Kumar Virwani, Masatoshi Ishii, Pritish Narayanan, Alessandro
  Fumarola, Lucas~L. Sanches, Irem Boybat, Manuel Le~Gallo, Kibong Moon, Jiyoo
  Woo, Hyunsang Hwang, and Yusuf Leblebici.
\newblock Neuromorphic computing using non-volatile memory.
\newblock \emph{Advances in Physics: X}, 2\penalty0 (1):\penalty0 89--124,
  2017.

\bibitem[Cappe et~al.(2012)Cappe, Rouiller, and Barone]{Cappe:2012}
Celine Cappe, Eric~M. Rouiller, and Pascal Barone.
\newblock \emph{Cortical and Thalamic Pathways for Multisensory and
  Sensorimotor Interplay}, chapter~2.
\newblock CRC Press/Taylor \& Francis, Boca Raton, Florida, USA, 2012.

\bibitem[Cooke and Bear(2015)]{Cooke:2015}
Sam~F. Cooke and Mark~F. Bear.
\newblock Visual recognition memory: a view from v1.
\newblock \emph{Current Opinion in Neurobiology}, 35:\penalty0 57--65, 2015.

\bibitem[Cox and Narayanan(2019)]{Cox:2019}
Charles~E. Cox and Pritish Narayanan.
\newblock Inc.
\newblock \emph{arXiv}, 2019.

\bibitem[Crick and Koch(2003)]{Crick:2003}
Francis Crick and Christof Koch.
\newblock A framework for consciousness.
\newblock \emph{Nat Neurosci}, 6\penalty0 (2):\penalty0 119--126, 2003.
\newblock 10.1038/nn0203-119.

\bibitem[Davies et~al.(2018)Davies, Srinivasa, Lin, Chinya, Cao, Choday, Dimou,
  Joshi, Imam, Jain, Liao, Lin, Lines, Liu, Mathaikutty, McCoy, Paul, Tse,
  Venkataramanan, Weng, Wild, Yang, and Wang]{Davies:2018}
M.~Davies, N.~Srinivasa, T.~H. Lin, G.~Chinya, Y.~Cao, S.~H. Choday, G.~Dimou,
  P.~Joshi, N.~Imam, S.~Jain, Y.~Liao, C.~K. Lin, A.~Lines, R.~Liu,
  D.~Mathaikutty, S.~McCoy, A.~Paul, J.~Tse, G.~Venkataramanan, Y.~H. Weng,
  A.~Wild, Y.~Yang, and H.~Wang.
\newblock Loihi: A neuromorphic manycore processor with on-chip learning.
\newblock \emph{IEEE Micro}, 38\penalty0 (1):\penalty0 82--99, 2018.

\bibitem[DiCarlo et~al.(2012)DiCarlo, Zoccolan, and Rust]{DiCarlo:2012}
James J DiCarlo, Davide Zoccolan, and Nicole C Rust.
\newblock How does the brain solve visual object recognition?
\newblock \emph{Neuron}, 73\penalty0 (3):\penalty0 415--434, 2012.

\bibitem[D’Souza and Burkhalter(2017)]{DSouza:2017}
Rinaldo~D. D’Souza and Andreas Burkhalter.
\newblock A laminar organization for selective cortico-cortical communication.
\newblock \emph{Frontiers in Neuroanatomy}, 11\penalty0 (71), 2017.

\bibitem[Elman(1990)]{Elman:1990}
Jeffrey~L. Elman.
\newblock Finding structure in time.
\newblock \emph{Cognitive Sci.}, 14:\penalty0 179--211, 1990.

\bibitem[Felleman and Van~Essen(1991)]{Felleman:1991}
Daniel~J. Felleman and David~C. Van~Essen.
\newblock Distributed hierarchical processing in the primate cerebral cortex.
\newblock \emph{Cereb. Cortex}, 1:\penalty0 1--47, 1991.

\bibitem[French(1999)]{French:1999}
Robert~M. French.
\newblock Catastrophic forgetting in connectionist networks.
\newblock \emph{Trends Cognit. Sci.}, 3:\penalty0 128--135, 1999.

\bibitem[Goodfellow et~al.(2016)Goodfellow, Bengio, and
  Courville]{Goodfellow:2016}
Ian Goodfellow, Yoshua Bengio, and Aaron Courville.
\newblock \emph{Deep Learning}.
\newblock MIT Press, 2016.

\bibitem[Gosmann and Eliasmith(2019)]{Gosmann:2019}
Jan Gosmann and Chris Eliasmith.
\newblock Vector-derived transformation binding: An improved binding operation
  for deep symbol-like processing in neural networks.
\newblock \emph{Neural Computation}, 31\penalty0 (5):\penalty0 849--869, 2019.

\bibitem[Guy et~al.(1999)Guy, ffytche, Brovelli, and Chumillas]{Guy:1999}
C.~N. Guy, D.~H. ffytche, A.~Brovelli, and J.~Chumillas.
\newblock fmri and eeg responses to periodic visual stimulation.
\newblock \emph{NeuroImage}, 10\penalty0 (2):\penalty0 125--148, 1999.

\bibitem[Haile et~al.(2019)Haile, Bohon, Romero, and Conway]{Haile:2019}
Theodros~M. Haile, Kaitlin~S. Bohon, Maria~C. Romero, and Bevil~R. Conway.
\newblock Visual stimulus-driven functional organization of macaque prefrontal
  cortex.
\newblock \emph{NeuroImage}, 188:\penalty0 427--444, 2019.

\bibitem[Harnad(1990)]{Harnad:1990}
Stevan Harnad.
\newblock The symbol grounding problem.
\newblock \emph{Physica D: Nonlinear Phenomena}, 42\penalty0 (1):\penalty0
  335--346, 1990.

\bibitem[Hawkins and Blakeslee(2004)]{Hawkins:2004}
Jeff Hawkins and Sandra Blakeslee.
\newblock \emph{On Intelligence}.
\newblock Times Books, Henry Holt \& Co., New York, NY, 2004.

\bibitem[Hawkins et~al.(2011)Hawkins, Ahmad, and Dubinsky]{Hawkins:2011}
Jeff Hawkins, Subutai Ahmad, and Donna Dubinsky.
\newblock Hierarchical temporal memory including htm cortical learning
  algorithms, version 0.2.1, 2011.
\newblock URL \url{numenta.com/assets/pdf/whitepapers}.

\bibitem[Hawkins et~al.(2017)Hawkins, Ahmad, and Cui]{Hawkins:2017}
Jeff Hawkins, Subutai Ahmad, and Yuwei Cui.
\newblock A theory of how columns in the neocortex enable learning the
  structure of the world.
\newblock \emph{Front. Neural Circuits}, 11, 2017.

\bibitem[Hebb(1949)]{Hebb:1949}
Donald~O. Hebb.
\newblock \emph{The Organization of Behavior: A Neuropsychological Theory}.
\newblock Wiley, N.Y., 1949.

\bibitem[Heeger(2017)]{Heeger:2017}
David~J. Heeger.
\newblock Theory of cortical function.
\newblock \emph{Proceedings of the National Academy of Sciences}, 114\penalty0
  (8):\penalty0 1773--1782, 2017.

\bibitem[Hubel and Wiesel(1959)]{Hubel:1959}
D.~H. Hubel and T.~N. Wiesel.
\newblock Receptive fields of single neurones in the cat's striate cortex.
\newblock \emph{The Journal of physiology}, 148\penalty0 (3):\penalty0
  574--591, 1959.

\bibitem[Kahnt(2017)]{Kahnt:2017}
Thorsten Kahnt.
\newblock A decade of decoding reward-related fmri signals and where we go from
  here.
\newblock \emph{NeuroImage}, 2017.

\bibitem[Kanerva(1988)]{Kanerva:1988}
Pentti Kanerva.
\newblock \emph{Sparse Distributed Memory}.
\newblock MIT Press, Cambridge, MA, 1988.

\bibitem[Laughlin and Sejnowski(2003)]{Laughlin:2003}
Simon~B. Laughlin and Terrence~J. Sejnowski.
\newblock Communication in neuronal networks.
\newblock \emph{Science}, 301\penalty0 (5641):\penalty0 1870, 2003.

\bibitem[LeCun et~al.(2015)LeCun, Bengio, and Hinton]{LeCun:2015}
Yann LeCun, Yoshua Bengio, and Geoffrey Hinton.
\newblock Deep learning.
\newblock \emph{Nature}, 521:\penalty0 436, 2015.

\bibitem[London and Hausser(2005)]{London:2005}
Michael London and Michael Hausser.
\newblock Dendritic computation.
\newblock \emph{Annu. Rev. Neurosci.}, 28:\penalty0 503--532, 2005.

\bibitem[Maass(2000)]{Maass:2000}
Wolfgang Maass.
\newblock On the computational power of winner-take-all.
\newblock \emph{Neural Computation}, 12\penalty0 (11):\penalty0 2519--2535,
  2000.

\bibitem[Markram et~al.(1997)Markram, Lubke, Frotscher, and
  Sakmann]{Markram:1997}
Henry Markram, Joachim Lubke, Michael Frotscher, and Bert Sakmann.
\newblock Regulation of synaptic efficacy by coincidence of postsynaptic aps
  and epsps.
\newblock \emph{Science}, 275:\penalty0 213--215, 1997.

\bibitem[Mattheij(2016)]{Mattheij:2016}
Jacques Mattheij.
\newblock Another way of looking at lee sedol vs alphago (blog), 2016.
\newblock URL
  \url{https://jacquesmattheij.com/another-way-of-looking-at-lee-sedol-vs-alphago/}.

\bibitem[McCulloch and Pitts(1943)]{MCP:1943}
Warren.~S. McCulloch and Walter.~H. Pitts.
\newblock A logical calculus of the ideas immanent in nervous activity.
\newblock \emph{B. Math. Biophys.}, 5:\penalty0 115--133, 1943.

\bibitem[Meijer et~al.(2019)Meijer, Mertens, Pennartz, Olcese, and
  Lansink]{Meijer:2019}
Guido~T. Meijer, Paul E.~C. Mertens, Cyriel M.~A. Pennartz, Umberto Olcese, and
  Carien~S. Lansink.
\newblock The circuit architecture of cortical multisensory processing:
  Distinct functions jointly operating within a common anatomical network.
\newblock \emph{Progress in Neurobiology}, 174:\penalty0 1--15, 2019.

\bibitem[Mountcastle(1957)]{Mountcastle:1957}
Vernon~B. Mountcastle.
\newblock Modality and topographic properties of single neurons of cat's
  somatic sensory cortex.
\newblock \emph{J. Neurophysiol.}, 20:\penalty0 408--434, 1957.

\bibitem[Mountcastle(1997)]{Mountcastle:1997}
Vernon~B. Mountcastle.
\newblock The columnar organization of the neocortex.
\newblock \emph{Brain}, 120:\penalty0 701--22, 1997.

\bibitem[Nakajima and Halassa(2017)]{Nakajima:2017}
Miho Nakajima and Michael~M. Halassa.
\newblock Thalamic control of functional cortical connectivity.
\newblock \emph{Current Opinion in Neurobiology}, 44\penalty0 (Supplement
  C):\penalty0 127--131, 2017.

\bibitem[O'Doherty et~al.(2017)O'Doherty, Cockburn, and Pauli]{ODoherty:2017}
John~P. O'Doherty, Jeffrey Cockburn, and Wolfgang~M. Pauli.
\newblock Learning, reward, and decision making.
\newblock \emph{Annual review of psychology}, 68\penalty0 (1):\penalty0
  73--100, 2017.

\bibitem[Oja(1982)]{Oja:1982}
Erkki Oja.
\newblock Simplified neuron model as a principal component analyzer.
\newblock \emph{J. Math. Biol.}, 15:\penalty0 267--273, 1982.

\bibitem[Olshausen and Field(1996)]{Olshausen:1996}
Bruno~A. Olshausen and David~J. Field.
\newblock Emergence of simple-cell receptive field properties by learning a
  sparse code for natural images.
\newblock \emph{Nature}, 381:\penalty0 607, 1996.

\bibitem[O'Reilly et~al.(2013)O'Reilly, Wyatte, Herd, Mingus, and
  Jilk]{OReilly:2013}
Randall~C. O'Reilly, Dean Wyatte, Seth Herd, Brian Mingus, and David Jilk.
\newblock Recurrent processing during object recognition.
\newblock \emph{Frontiers in Psychology}, 4\penalty0 (124), 2013.

\bibitem[O'Reilly et~al.(2014)O'Reilly, Wyatte, and Rohrlich]{OReilly:2014}
Randall~C. O'Reilly, Dean Wyatte, and John Rohrlich.
\newblock Learning through time in the thalamocortical loops, 2014.
\newblock URL \url{https://arxiv.org/abs/1407.3432}.

\bibitem[Pehlevan et~al.(2018)Pehlevan, Sengupta, and
  Chklovskii]{Pehlevan:2018}
Cengiz Pehlevan, Anirvan~M. Sengupta, and Dmitri~B. Chklovskii.
\newblock Why do similarity matching objectives lead to hebbian/anti-hebbian
  networks?
\newblock \emph{Neural Computation}, 30\penalty0 (1):\penalty0 84--124, 2018.

\bibitem[Perez-Orive et~al.(2002)Perez-Orive, Mazor, Turner, Cassenaer, Wilson,
  and Laurent]{Perez-Orive:2002}
Javier Perez-Orive, Ofer Mazor, Glenn~C. Turner, Stijn Cassenaer, Rachel~I.
  Wilson, and Gilles Laurent.
\newblock Oscillations and sparsening of odor representations in the mushroom
  body.
\newblock \emph{Science}, 297\penalty0 (5580):\penalty0 359, 2002.

\bibitem[Plate(1995)]{Plate:1995}
T.~A. Plate.
\newblock Holographic reduced representations.
\newblock \emph{IEEE Transactions on Neural Networks}, 6\penalty0 (3):\penalty0
  623--641, 1995.

\bibitem[Purves et~al.(2018)Purves, Augustine, Fitzpatrick, Hall, LaMantia,
  Mooney, Platt, and White]{Purves:2018}
Dale Purves, George~J Augustine, David Fitzpatrick, William~C. Hall,
  Anthony-Samuel LaMantia, Richard~D. Mooney, Michael~L. Platt, and Leonard~E.
  White.
\newblock \emph{Neuroscience}.
\newblock Sinauer, Sunderland, Massachusetts, USA, 6 edition, 2018.

\bibitem[Quiroga et~al.(2005)Quiroga, Reddy, Kreiman, Koch, and
  Fried]{Quiroga:2005}
R.~Quian Quiroga, L.~Reddy, G.~Kreiman, C.~Koch, and I.~Fried.
\newblock Invariant visual representation by single neurons in the human brain.
\newblock \emph{Nature}, 435:\penalty0 1102--1107, 2005.

\bibitem[Reynolds and Wickens(2002)]{Reynolds:2002}
John N.~J. Reynolds and Jeffery~R. Wickens.
\newblock Dopamine-dependent plasticity of corticostriatal synapses.
\newblock \emph{Neural Networks}, 15\penalty0 (4):\penalty0 507--521, 2002.

\bibitem[Riesenhuber and Poggio(1999)]{Riesenhuber:1999}
Maximilian Riesenhuber and Tomaso Poggio.
\newblock Hierarchical models of object recognition in cortex.
\newblock \emph{Nature Neuroscience}, 2\penalty0 (11):\penalty0 1019--1025,
  1999.

\bibitem[Rinkus(2010)]{Rinkus:2010}
Gerard Rinkus.
\newblock A cortical sparse distributed coding model linking mini- and
  macrocolumn-scale functionality.
\newblock \emph{Front. Neuroanat.}, 4, 2010.

\bibitem[Rinkus(1993)]{Rinkus:1993}
Gerard~J. Rinkus.
\newblock Context-sensitive spatio-temporal memory, 1993.
\newblock Technical report.

\bibitem[Sakagami et~al.(2006)Sakagami, Pan, and Uttl]{Sakagami:2006}
Masamichi Sakagami, Xiaochuan Pan, and Bob Uttl.
\newblock Behavioral inhibition and prefrontal cortex in decision-making.
\newblock \emph{Neural Networks}, 19\penalty0 (8):\penalty0 1255--1265, 2006.

\bibitem[Sanger(1989)]{Sanger:1989}
Terence~D. Sanger.
\newblock Optimal unsupervised learning in a single-layer linear feedforward
  neural network.
\newblock \emph{Neural Networks}, 2:\penalty0 459--473, 1989.

\bibitem[Shouval et~al.(2010)Shouval, Wang, and Wittenberg]{Shouval:2010}
Harel Shouval, Samuel Wang, and Gayle Wittenberg.
\newblock Spike timing dependent plasticity: A consequence of more fundamental
  learning rules.
\newblock \emph{Frontiers in Computational Neuroscience}, 4\penalty0 (19),
  2010.

\bibitem[Silver et~al.(2016)Silver, Huang, Maddison, Guez, Sifre, van~den
  Driessche, Schrittwieser, Antonoglou, Panneershelvam, Lanctot, Dieleman,
  Grewe, Nham, Kalchbrenner, Sutskever, Lillicrap, Leach, Kavukcuoglu, Graepel,
  and Hassabis]{Silver:2016}
David Silver, Aja Huang, Chris~J. Maddison, Arthur Guez, Laurent Sifre, George
  van~den Driessche, Julian Schrittwieser, Ioannis Antonoglou, Veda
  Panneershelvam, Marc Lanctot, Sander Dieleman, Dominik Grewe, John Nham, Nal
  Kalchbrenner, Ilya Sutskever, Timothy Lillicrap, Madeleine Leach, Koray
  Kavukcuoglu, Thore Graepel, and Demis Hassabis.
\newblock Mastering the game of go with deep neural networks and tree search.
\newblock \emph{Nature}, 529:\penalty0 484, 2016.

\bibitem[Thakur et~al.(2018)Thakur, Molin, Cauwenberghs, Indiveri, Kumar, Qiao,
  Schemmel, Wang, Chicca, Olson~Hasler, Seo, Yu, Cao, van Schaik, and
  Etienne-Cummings]{Thakur:2018}
Chetan~Singh Thakur, Jamal~Lottier Molin, Gert Cauwenberghs, Giacomo Indiveri,
  Kundan Kumar, Ning Qiao, Johannes Schemmel, Runchun Wang, Elisabetta Chicca,
  Jennifer Olson~Hasler, Jae-sun Seo, Shimeng Yu, Yu~Cao, André van Schaik,
  and Ralph Etienne-Cummings.
\newblock Large-scale neuromorphic spiking array processors: A quest to mimic
  the brain.
\newblock \emph{Frontiers in Neuroscience}, 12\penalty0 (891), 2018.

\bibitem[Tsunoda et~al.(2001)Tsunoda, Yamane, Nishizaki, and
  Tanifuji]{Tsunoda:2001}
Kazushige Tsunoda, Yukako Yamane, Makoto Nishizaki, and Manabu Tanifuji.
\newblock Complex objects are represented in macaque inferotemporal cortex by
  the combination of feature columns.
\newblock \emph{Nat Neurosci}, 4\penalty0 (8):\penalty0 832--838, 2001.
\newblock 10.1038/90547.

\bibitem[Werbos(1982)]{Werbos:1982}
Paul~J Werbos.
\newblock Application and advances in nonlinear sensitivity analysis, 1982.

\bibitem[Wunderlich et~al.(2019)Wunderlich, Kungl, Müller, Hartel, Stradmann,
  Aamir, Grübl, Heimbrecht, Schreiber, Stöckel, Pehle, Billaudelle, Kiene,
  Mauch, Schemmel, Meier, and Petrovici]{Wunderlich:2019}
Timo Wunderlich, Akos~F. Kungl, Eric Müller, Andreas Hartel, Yannik Stradmann,
  Syed~Ahmed Aamir, Andreas Grübl, Arthur Heimbrecht, Korbinian Schreiber,
  David Stöckel, Christian Pehle, Sebastian Billaudelle, Gerd Kiene, Christian
  Mauch, Johannes Schemmel, Karlheinz Meier, and Mihai~A. Petrovici.
\newblock Demonstrating advantages of neuromorphic computation: A pilot study.
\newblock \emph{Frontiers in Neuroscience}, 13\penalty0 (260), 2019.

\bibitem[Yan et~al.(2019)Yan, Kappel, Neumaerker, Partzsch, Vogginger,
  Hoeppner, Furber, Maass, Legenstein, and Mayr]{YYan:2019}
Y.~Yan, D.~Kappel, F.~Neumaerker, J.~Partzsch, B.~Vogginger, S.~Hoeppner,
  S.~Furber, W.~Maass, R.~Legenstein, and C.~Mayr.
\newblock Efficient reward-based structural plasticity on a spinnaker 2
  prototype.
\newblock \emph{IEEE Transactions on Biomedical Circuits and Systems}, pages
  1--1, 2019.

\bibitem[Yildirim et~al.(2019)Yildirim, Wu, Kanwisher, and
  Tenenbaum]{Yildirim:2019}
Ilker Yildirim, Jiajun Wu, Nancy Kanwisher, and Joshua Tenenbaum.
\newblock An integrative computational architecture for object-driven cortex.
\newblock \emph{Current Opinion in Neurobiology}, 55:\penalty0 73--81, 2019.

\bibitem[Zipoli~Caiani and Ferretti(2017)]{Caiani:2017}
Silvano Zipoli~Caiani and Gabriele Ferretti.
\newblock Semantic and pragmatic integration in vision for action.
\newblock \emph{Consciousness and Cognition}, 48:\penalty0 40--54, 2017.

\bibitem[Zong et~al.(2017)Zong, Ge, and Gu]{Zong:2017}
Ziliang Zong, Rong Ge, and Qijun Gu.
\newblock Marcher: A heterogeneous system supporting energy-aware high
  performance computing and big data analytics.
\newblock \emph{Big Data Research}, 8:\penalty0 27--38, 2017.

\end{thebibliography}

\end{document}